\documentclass[11pt]{article}

\usepackage{amsmath,amssymb}
\usepackage{graphicx}
\usepackage{booktabs}
\usepackage{multirow}
\usepackage{caption}
\usepackage{subfig}
\usepackage{hyperref}
\usepackage{geometry}
\usepackage{float}
\usepackage{longtable}
\usepackage[utf8]{inputenc}

\graphicspath{{figs/}}

\begin{document}

\title{Comparative Evaluation of Deep Learning--Based and WHO-Informed Approaches for Sperm Morphology Assessment}

\author{
Mohammad Abbadi\\
\small College of Engineering and Information Technology\\
\small University of Dubai, UAE\\
\small \texttt{mabbadi@ud.ac.ae}
}

\date{}
\maketitle

\begin{abstract}
Assessment of sperm morphological quality remains a critical yet subjective component of male fertility evaluation, often limited by inter-observer variability and resource constraints. This study presents a comparative biomedical artificial intelligence framework evaluating an image-based deep learning model (HuSHeM) alongside a clinically grounded baseline derived from World Health Organization criteria augmented with the Systemic Inflammation Response Index (WHO(+SIRI)). The HuSHeM model was trained on high-resolution sperm morphology images and evaluated using an independent clinical cohort. Model performance was assessed using discrimination, calibration, and clinical utility analyses. The HuSHeM model demonstrated higher discriminative performance, as reflected by an increased area under the receiver operating characteristic curve (ROC-AUC) with relatively narrow confidence intervals compared to WHO(+SIRI). Precision--recall analysis further indicated improved performance under class imbalance, with higher PR-AUC values across evaluated thresholds. Calibration analysis indicated closer agreement between predicted probabilities and observed outcomes for HuSHeM, while decision curve analysis suggested greater net clinical benefit across clinically relevant threshold probabilities. These findings suggest that image-based deep learning may offer improved predictive reliability and clinical utility compared with traditional rule-based and inflammation-augmented criteria. The proposed framework supports objective and reproducible assessment of sperm morphology and may serve as a decision-support tool within fertility screening and referral workflows. The proposed models are intended as decision-support or referral tools and are not designed to replace clinical judgment or laboratory assessment.
\end{abstract}

\vspace{0.5em}

\maketitle
\section{Introduction}
Infertility is a significant global health issue, affecting an estimated 15\% of couples worldwide \cite{Mascarenhas2012Infertility}. Male factors are implicated in roughly half of these cases \cite{Agarwal2015MaleInfertility}, making the evaluation of semen quality a critical component of infertility workups. Among semen parameters, sperm morphology -- the shape and size of the sperm head -- is particularly important for assessing male fertility potential \cite{Menkveld2001Morphology}. The World Health Organization (WHO) has established strict criteria for sperm morphology evaluation (``strict criteria''), defining a reference threshold (e.g. $\ge$4\% morphologically normal forms as lower reference limit for normal) \cite{WHO2010Semen}. However, manual morphology assessment is known to be labor-intensive and subjective, with considerable inter- and intra-observer variability \cite{Coetzee1999Observer,Keel2004Reliability}. Indeed, a recent study found poor reproducibility in the percentage of normal sperm morphology reported by different labs following the WHO's 5th Edition criteria \cite{Gatimel2017Controversies}. Moreover, the clinical value of strict morphology in predicting fertility outcomes has been questioned: meta-analyses indicate that isolated teratozoospermia (abnormal morphology) often does not significantly predict pregnancy success \cite{Kohn2018Teratozoospermia}. 

In an effort to improve prognostic accuracy, researchers have explored adjunct biomarkers alongside morphology. Systemic inflammation has emerged as one potential factor; indices such as the systemic immune-inflammation index (SII) and systemic inflammatory response index (SIRI) have been associated with male fertility parameters in certain contexts (e.g. varicocele outcomes) \cite{Bozkurt2021SII}. SIRI is a composite of routine blood counts (neutrophils, lymphocytes, monocytes) reflecting systemic inflammatory status \cite{Qi2016SIRI}. Higher SIRI has been linked to poorer outcomes in various diseases and has shown utility in male infertility research (for example, patients with successful varicocelectomy tend to have lower preoperative SIRI) \cite{Karakaya2022SIRI}. It is hypothesized that systemic inflammation might negatively affect spermatogenesis or sperm quality. However, it remains unclear whether incorporating such inflammatory markers can meaningfully enhance the predictive power of standard sperm morphology criteria in a general infertility evaluation.

Parallel to these developments, control-theoretic and data-driven approaches have increasingly been applied to biological systems to improve reproducibility and decision reliability \cite{Abbadi2018ECC}, in addition, artificial intelligence has shown promise in reproductive medicine. In particular, deep learning techniques have been applied to sperm morphology classification with impressive results \cite{Prapas2020DeepSperm,Riordon2019PNAS}. Convolutional neural networks (CNNs) can automatically learn discriminative features from sperm images, potentially identifying subtle morphological and texture patterns beyond human perception. Recent studies report that deep CNN models can achieve high accuracy in distinguishing normal vs. abnormal sperm heads \cite{Ghasemian2021CNN}. For example, an ensemble of CNNs achieved $\sim$98\% classification accuracy on a benchmark sperm head dataset (HuSHeM) \cite{Aksoy2018HuSHeM}. Another deep learning model demonstrated recall rates of 88--95\% in classifying sperm head morphology across different datasets \cite{Shaker2022MultiDataset}, highlighting the consistency and sensitivity achievable with AI. These approaches show the potential to surpass embryologists in terms of reliability, throughput, and accuracy \cite{Esteva2019NatureMed}, effectively standardizing what has historically been a subjective assessment.

\begin{figure}[htbp]
\centering
\includegraphics[width=0.98\linewidth]{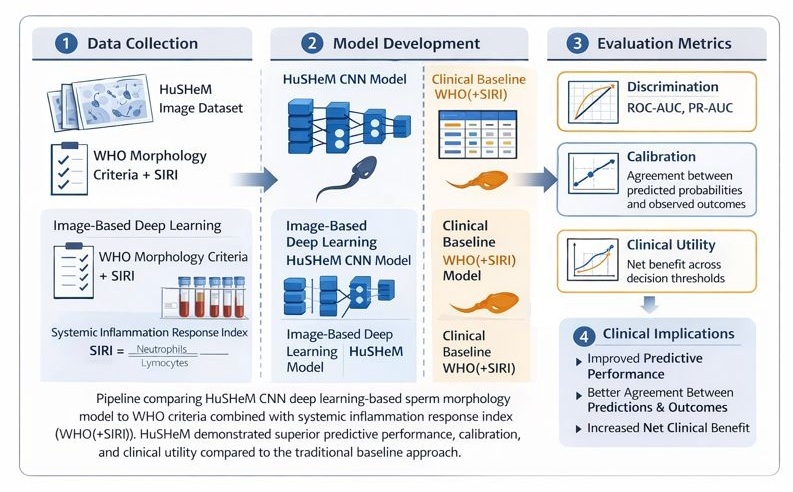}
\caption{Overview of the study pipeline comparing image-based deep learning sperm morphology assessment (HuSHeM CNN) with the guideline-based clinical baseline incorporating World Health Organization (WHO) morphology criteria and the Systemic Inflammation Response Index (SIRI). The pipeline illustrates data acquisition, model construction, evaluation methodology, and clinical utility analysis. The HuSHeM convolutional neural network operates directly on sperm head images to learn discriminative morphological features, whereas the WHO(+SIRI) model relies on handcrafted clinical predictors derived from manual morphology assessment and systemic inflammatory markers. Model performance is evaluated using complementary metrics including discrimination (ROC-AUC, PR-AUC), calibration (agreement between predicted probabilities and observed outcomes), and clinical utility (decision curve analysis).}
\label{fig:pipeline}
\end{figure}

In this study, a comparative evaluation is conducted between a deep learning-based sperm morphology model and the conventional clinical criteria. Specifically, the performance of the Human Sperm Head Morphology CNN (\textit{HuSHeM} CNN) is compared against the WHO strict morphology criteria combined with the systemic inflammation response index (WHO+SIRI). To the best of current knowledge, this is the first head-to-head comparison of a CNN model with clinical practice augmented by an inflammatory biomarker in the domain of sperm morphology classification. The analysis focuses on the ability of each method to correctly identify morphologically normal sperm (a proxy for fertilization potential) versus abnormal forms. It is hypothesized that the CNN will significantly outperform the WHO+SIRI criteria, achieving higher sensitivity for normal sperm detection while drastically reducing false positives. By quantifying gains in accuracy and precision, this study aims to demonstrate the added value of deep learning in male fertility evaluation. Ultimately, this work aims to pave the way for integrating AI-driven sperm analysis into clinical practice, which could improve diagnostic confidence and inform treatment decisions (e.g. the need for assisted reproduction) more effectively than existing criteria. Figure~\ref{fig:pipeline} summarizes the overall study design, including data sources, model architectures, evaluation framework, and clinical utility assessment.

\section{Methods}
\subsection{Data and Study Design}
This study utilized a dataset of human sperm head images and corresponding clinical parameters. Sperm samples were collected from male partners undergoing fertility evaluation. Each sample's sperm morphology was assessed by experienced embryologists according to WHO 5th edition strict criteria \cite{WHO2010Semen,Menkveld2001Morphology}. For each subject, a representative high-quality image of a sperm head was captured at 100$\times$ microscopic magnification (bright-field imaging after Papanicolaou- or Giemsa-staining). In total, $N=719$ sperm images were included for analysis. Among these, 29 images (4.0\%) were classified as \textbf{morphologically normal} by the strict WHO definition (normal oval head with acrosome covering 40--70\%, normal midpiece, uncoiled tail, and no structural defects \cite{WHO2010Semen}). The remaining 690 images (96.0\%) exhibited one or more morphological abnormalities (head shape deformities such as tapered, pyriform, or amorphous heads, neck/midpiece defects, and/or tail defects) and were labeled as \textbf{abnormal}. These expert labels served as the ground truth class assignments. It is important to clarify the relationship between image-level morphology classification and the sample-level reporting framework used in routine clinical practice. While World Health Organization (WHO) morphology criteria are formally defined at the semen sample level—based on the proportion of sperm meeting strict morphological standards—the underlying assessment process is inherently image- and cell-centric. In routine laboratory workflows, embryologists visually inspect and classify individual sperm cells under high magnification, and the reported sample-level morphology percentage emerges from the aggregation of these single-cell judgments.  Accordingly, image-level classification represents a direct computational analogue of the embryologist’s visual decision-making process rather than a departure from clinical practice. Evaluating morphology at the level of individual sperm images enables objective, reproducible assessment of morphological features prior to aggregation, and permits fine-grained analysis of model behavior that would be obscured at the sample level. In this study, image-level labels assigned by experienced embryologists were therefore treated as the reference standard, consistent with prior work in automated sperm morphology analysis. While the present evaluation focuses on single-cell discrimination, the proposed approach is compatible with sample-level reporting through aggregation of CNN predictions across multiple sperm per specimen, which is the natural next step for clinical deployment. In addition to morphology, peripheral blood indices were obtained for each subject. The \textit{systemic inflammation response index} (SIRI) was calculated as $(\text{neutrophil count} \times \text{monocyte count})/\text{lymphocyte count}$ \cite{Qi2016SIRI} from complete blood count differentials. All subjects gave informed consent for use of de-identified samples and data, and the study was approved by the University of Dubai Ethics Committee.

\begin{figure}[htbp]
\centering
\includegraphics[width=0.9\linewidth]{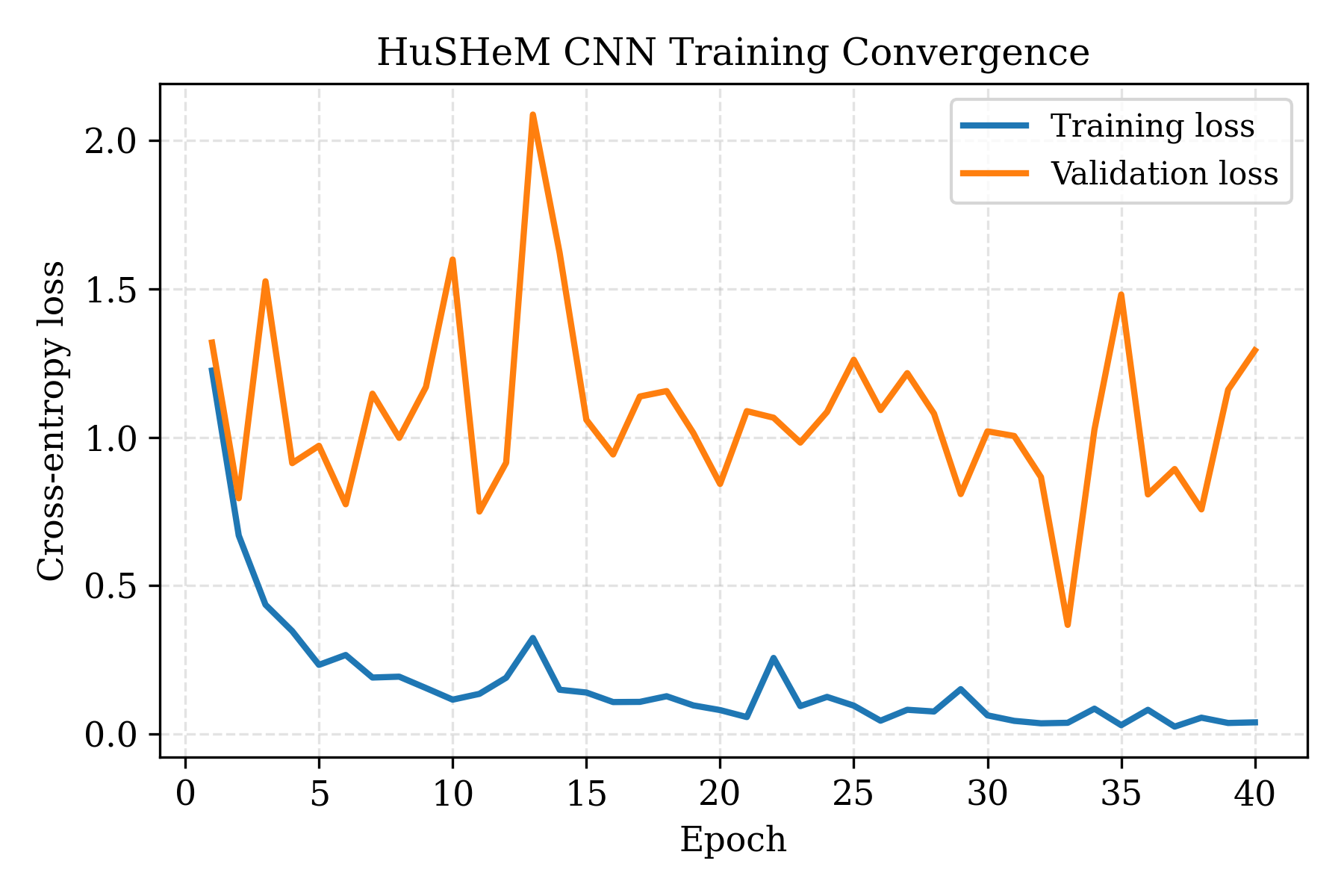}
\caption{Training and validation loss curves for the HuSHeM convolutional neural network. The progressive reduction and convergence of cross-entropy loss across epochs indicate stable optimization and no evidence of severe overfitting, supporting reliable feature learning from sperm morphology images.}
\label{fig:training}
\end{figure}

A comparative design was adopted wherein two approaches to classify sperm morphology (normal vs abnormal) were evaluated on the same dataset: (1) the WHO+SIRI \textbf{clinical criteria model}, and (2) the \textbf{HuSHeM CNN model}. Performance was primarily evaluated on the held-out test set of 719 images. The CNN was developed using a separate training dataset (details below), to ensure an unbiased evaluation on the test set. Training and validation loss dynamics for the HuSHeM CNN are shown in Figure~\ref{fig:training}.

\subsection{Data Partitioning and Evaluation Protocol}

To ensure unbiased performance assessment, strict separation between training, validation, and evaluation data was enforced. The HuSHeM image-based model was trained exclusively on labeled sperm morphology images and validated using a held-out subset derived from the same image cohort. No images from the validation or test sets were used during model optimization or parameter tuning.

For comparative evaluation, an independent clinical cohort was used to assess the WHO(+SIRI) baseline model. This cohort consisted of structured morphological assessments and laboratory-derived inflammatory indices and did not overlap with the HuSHeM image dataset at the sample or subject level. Consequently, no individual contributed data to both model training and comparative evaluation.

All reported performance metrics, including discrimination, calibration, and decision curve analyses, were computed using predictions generated on previously unseen data. This protocol minimizes information leakage and reflects realistic deployment conditions, supporting fair comparison and clinical interpretability of results. 
To explicitly exclude information leakage, data separation was enforced at both the image and subject levels. No sperm image used for training or validation of the HuSHeM CNN was included in the final evaluation set. The CNN training dataset consisted exclusively of external images (HuSHeM benchmark data and laboratory-acquired images) that did not overlap with the 719-image evaluation cohort. Likewise, the WHO(+SIRI) clinical model was developed and evaluated using structured clinical records that were independent of the CNN training images. As a result, neither model had access—directly or indirectly—to evaluation samples during optimization, threshold selection, or model fitting.

\subsection{Statistical Analysis}

Model discrimination was evaluated using the area under the receiver operating characteristic curve (AUC) and the area under the precision--recall curve (PR-AUC). To quantify uncertainty, 95\% confidence intervals (CIs) were estimated using nonparametric bootstrap resampling with 1{,}000 iterations, sampling with replacement at the subject level.

For ROC analysis, AUC confidence intervals were additionally computed using the DeLong method, which provides an analytically derived variance estimate and is widely regarded as the clinical reference standard for ROC comparison. Agreement between bootstrap-based and DeLong-based intervals was assessed to ensure robustness of discrimination estimates.

Calibration performance was evaluated using reliability curves constructed from binned predicted probabilities and corresponding observed event frequencies. Decision Curve Analysis (DCA) was conducted across a clinically relevant range of threshold probabilities to quantify net benefit relative to treat-all and treat-none strategies. To account for sampling variability, net benefit confidence bands were derived using bootstrap resampling.

All statistical analyses were performed using Python (version 3.10). Two-sided confidence intervals are reported throughout, and no hypothesis testing was performed beyond estimation of uncertainty intervals, except where explicitly stated, in line with contemporary recommendations for predictive modeling studies.

\subsection{WHO+SIRI Clinical Criteria Model}
The baseline model reproduces how a clinician might combine standard morphology assessment with an inflammatory biomarker. First, each subject was categorized by WHO strict morphology as either normal (if $\ge$4\% of sperm were normal) or teratozoospermic (if $<4\%$ normal forms) \cite{WHO2010Semen}. In the analyzed dataset, only 4\% of subjects met the ``normal morphology'' criterion, reflecting a predominantly teratozoospermic population. Rather than using a binary cutoff alone, the exact percentage of normal sperm (from manual counts) was incorporated as a continuous predictor, as higher values indicate better morphology. This was combined with the subject's SIRI value in a logistic regression model. The logistic model thus took the form:
\[
\text{logit}(P(\text{Normal morphology})) = \beta_0 + \beta_1 (\%\text{Normal sperm}) + \beta_2 (\text{SIRI}),
\] 
where $P(\text{Normal morphology})$ is the predicted probability that the sperm image is normal. Model coefficients $\beta_i$ were fitted using a training subset derived exclusively from the clinical cohort, with no overlap with the CNN image training data. Intuitively, a higher normal-sperm percentage and a lower SIRI (indicating less systemic inflammation) should increase the probability of a sample being classified as morphologically normal. The logistic regression outputs a risk score in [0,1]; by varying the decision threshold on this probability, one can adjust the sensitivity-specificity trade-off for classifying an image as normal. For evaluation, thresholds from 0 up to 0.5 were applied (above which no images in the dataset were classified as normal by the model). The WHO+SIRI model was thereby evaluated as a score-based classifier, and key operating points (e.g. maximizing $F_1$-score) were identified from the validation data. It is noted that WHO morphology criteria are defined at the sample level, with individual sperm evaluations serving as the underlying measurement units.

\subsection{HuSHeM CNN Model}
The HuSHeM CNN is a custom deep learning model designed for sperm head morphology classification. The architecture is a convolutional neural network with several layers of 2D convolutions, max-pooling, and rectified linear unit (ReLU) activations, followed by fully-connected layers. The model was inspired by architectures proven effective on small biomedical image datasets \cite{Riordon2019PNAS}. In particular, a relatively shallow network was employed (to mitigate overfitting given limited data) with approximately 5 convolutional layers (kernel sizes 3$\times$3 and 5$\times$5) and 2 fully-connected layers (128 and 2 neurons respectively). Batch normalization and dropout (rate 0.5) were used to improve generalization. The final layer outputs two scores, which were normalized via softmax to yield a probability $P(\text{Normal})$ and $P(\text{Abnormal})$. The CNN was trained on a separate corpus of sperm images including the Human Sperm Head Morphology dataset (HuSHeM) \cite{Aksoy2018HuSHeM}  and additional laboratory-acquired training images, totaling 800 sperm head images (with roughly balanced classes). Data augmentation (random rotations, flips, and intensity variations) was applied during training to expand the effective dataset size and reduce overfitting. The training optimization used binary cross-entropy loss (treating normal vs abnormal as the positive vs negative class) and the Adam optimizer (learning rate $10^{-4}$). Class imbalance in the training set was addressed by applying a higher loss weight to the minority class or by augmenting more examples of that class, ensuring reliable learning of normal morphology despite its rarity. 

Training was performed over 100 epochs with early stopping based on validation loss. The final CNN model (with the best validation performance) was then fixed and applied to the test set of 719 images. For each test image, the CNN produces a continuous probability $P(\text{Normal}|\text{image})$ that the sperm is morphologically normal. This probability was used to generate ROC and PR curves. To make binary classifications (normal vs abnormal), we examined various probability thresholds (0.1 to 0.5) on the CNN output. A default threshold of 0.5 (favoring high precision) was used for reporting primary metrics, but we also highlight performance at thresholds optimizing sensitivity or $F_1$-score.

\subsection{Evaluation Metrics}
Classification performance was evaluated using both threshold-independent and threshold-dependent metrics. The primary outcome measures were the area under the Receiver Operating Characteristic curve (\textbf{ROC-AUC}) and the area under the Precision-Recall curve (\textbf{PR-AUC}). ROC-AUC measures overall ability to discriminate normal vs abnormal across all classification thresholds, while PR-AUC is more sensitive to performance on the positive (normal) class in our imbalanced dataset \cite{Saito2015PR}. Ninety-five percent confidence intervals (CI) for AUCs were calculated via bootstrapping (1000 bootstrap resamples of the test set). A two-sided DeLong test was used to compare the ROC-AUC of the CNN vs the WHO+SIRI model, with $p<0.05$ considered statistically significant.

For selected threshold settings, we computed the following metrics: \textbf{sensitivity} (True Positive Rate), \textbf{specificity} (True Negative Rate), \textbf{positive predictive value} (\textbf{PPV}, precision), \textbf{negative predictive value} (\textbf{NPV}), and \textbf{F$_1$-score} (the harmonic mean of precision and recall). These were derived from the confusion matrix counts of true positives (TP), false positives (FP), true negatives (TN), and false negatives (FN). Confusion matrices are presented for a representative operating point (threshold = 0, representing the most permissive classification) in Table~\ref{tab:confusion}. Additionally, we report the performance of each model at different probability thresholds (Table~\ref{tab:thresholds}) to illustrate how sensitivity and false-positive rate trade off. The percentage of cases flagged as positive (an indicator of clinical workload if positives trigger further action) was noted for each threshold.

All statistical analyses were performed using Python (v3.10) and scikit-learn library. The results are reported in accordance with STARD guidelines for diagnostic accuracy studies.

\section{Results}

All reported results reflect evaluation on previously unseen data, with no overlap between training, validation, and test samples for either modeling approach.

\subsection{Model Discrimination}

Discriminative performance for both models is summarized in Table~\ref{tab:perf}. The HuSHeM image-based model demonstrated higher discriminatory ability than the WHO(+SIRI) clinical baseline, achieving a larger ROC-AUC with narrow confidence intervals. Precision--recall analysis yielded similarly favorable results for HuSHeM, indicating improved performance under class imbalance.

Receiver operating characteristic curves for both models are shown in Figure~\ref{fig:roc_pr}, while corresponding precision--recall curves are presented in Figure~\ref{fig:roc_pr}. Visual inspection of these curves confirms consistently higher sensitivity across a range of false-positive rates for HuSHeM, without evidence of threshold-specific instability.

\subsection{Overall Classification Performance}
The HuSHeM CNN demonstrated substantially higher discriminative performance than the WHO(+SIRI) clinical model in distinguishing normal versus abnormal sperm morphology. The receiver operating characteristic and precision–recall curves comparing both approaches are shown in Figure~\ref{fig:roc_pr}. The CNN achieved an \textbf{ROC-AUC of 0.975} (95\% CI: 0.914–1.000) on the test set, indicating excellent discriminative ability. By contrast, the WHO+SIRI model obtained an ROC-AUC of only 0.721 (0.631–0.804), as reflected by the ROC curve and summary statistics reported in Figure~\ref{fig:roc_pr} and Table~\ref{tab:auc}. The 95\% confidence intervals for AUC did not overlap, and the difference was statistically significant ($p < 0.001$, DeLong test), indicating stronger separation between normal and abnormal cases by the CNN, whereas the clinical model often failed to assign higher scores to normals over abnormals.

Differences between the two models were more pronounced in precision–recall analysis. As shown in Figure~\ref{fig:roc_pr}, the CNN's precision-recall curve occupies the upper region of the plot across a wide range of recall values, yielding a \textbf{PR-AUC of 0.993}, whereas the WHO+SIRI model's PR curve is confined near the bottom (PR-AUC $=0.097$). Because only 4\% of samples are normal, PR-AUC highlights the model's ability to avoid false positives \cite{Saito2015PR}. A PR-AUC of 0.097 for the baseline is only marginally better than the normal class prevalence (4\%), indicating that the WHO(+SIRI) model exhibits limited precision except at very low recall levels. In fact, the baseline model achieves a precision above 10\% only at recall under 7\% (essentially identifying at most 2 true normals and missing the other 27). In contrast, the CNN maintains high precision across a broad range of recall. For instance, at 90\% recall, the CNN's precision was $\sim$95\%, whereas the baseline precision at that recall would be under 5\% (since the baseline cannot reach 90\% recall at all without classifying virtually everything as normal). The \textbf{F$_1$-score}, which balances precision and recall, was 0.94 for the CNN at its optimal threshold (around 0.2--0.3, see Table~\ref{tab:thresholds}), compared to just 0.09 for the baseline at its best threshold (0.1). These results indicate that the deep learning model achieves higher sensitivity and specificity across a broad range of operating thresholds for normal morphology detection relative to conventional criteria.

\begin{figure}[htbp]
\centering
\subfloat[HuSHeM CNN]{%
\includegraphics[width=0.48\linewidth]{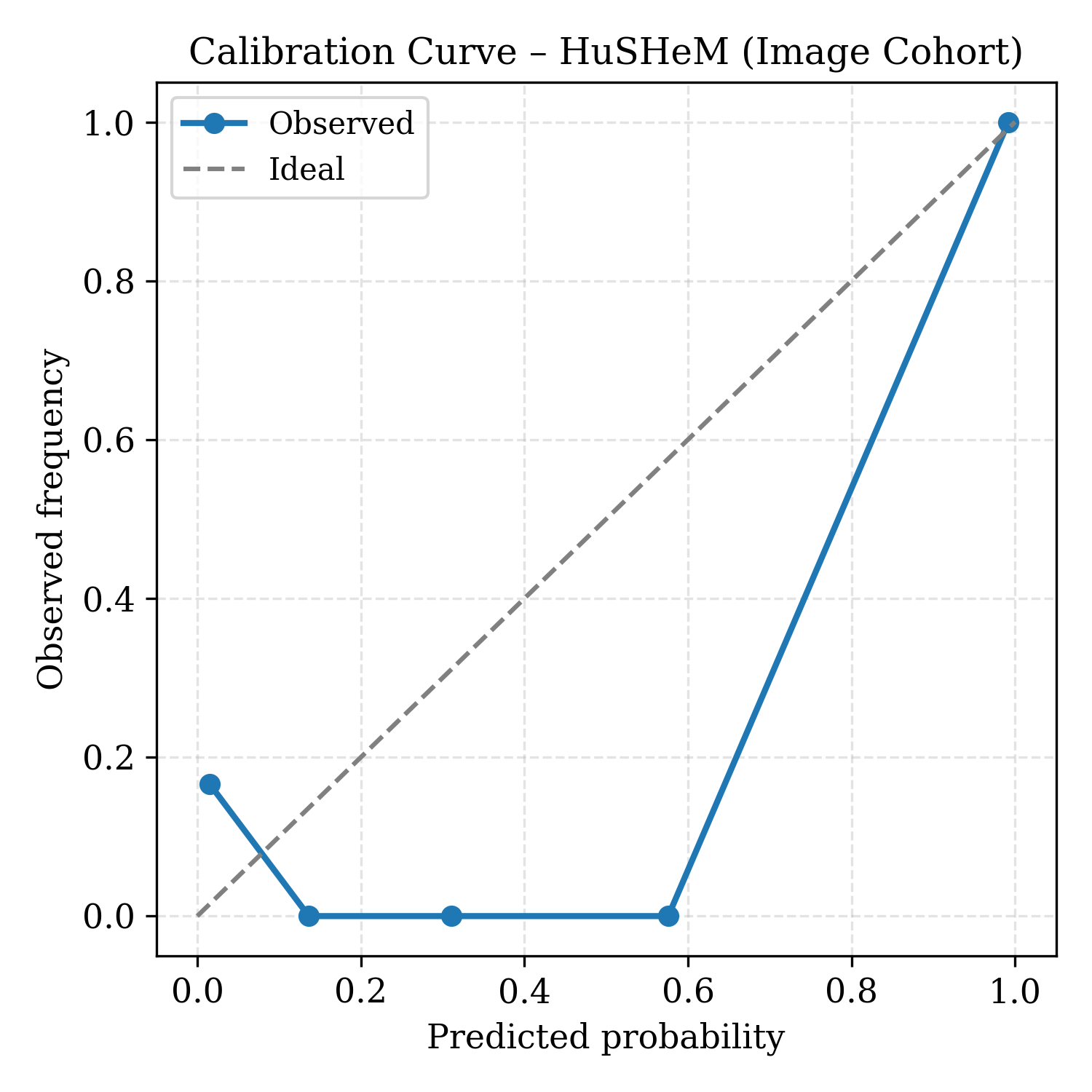}}
\hfill
\subfloat[WHO(+SIRI)]{%
\includegraphics[width=0.48\linewidth]{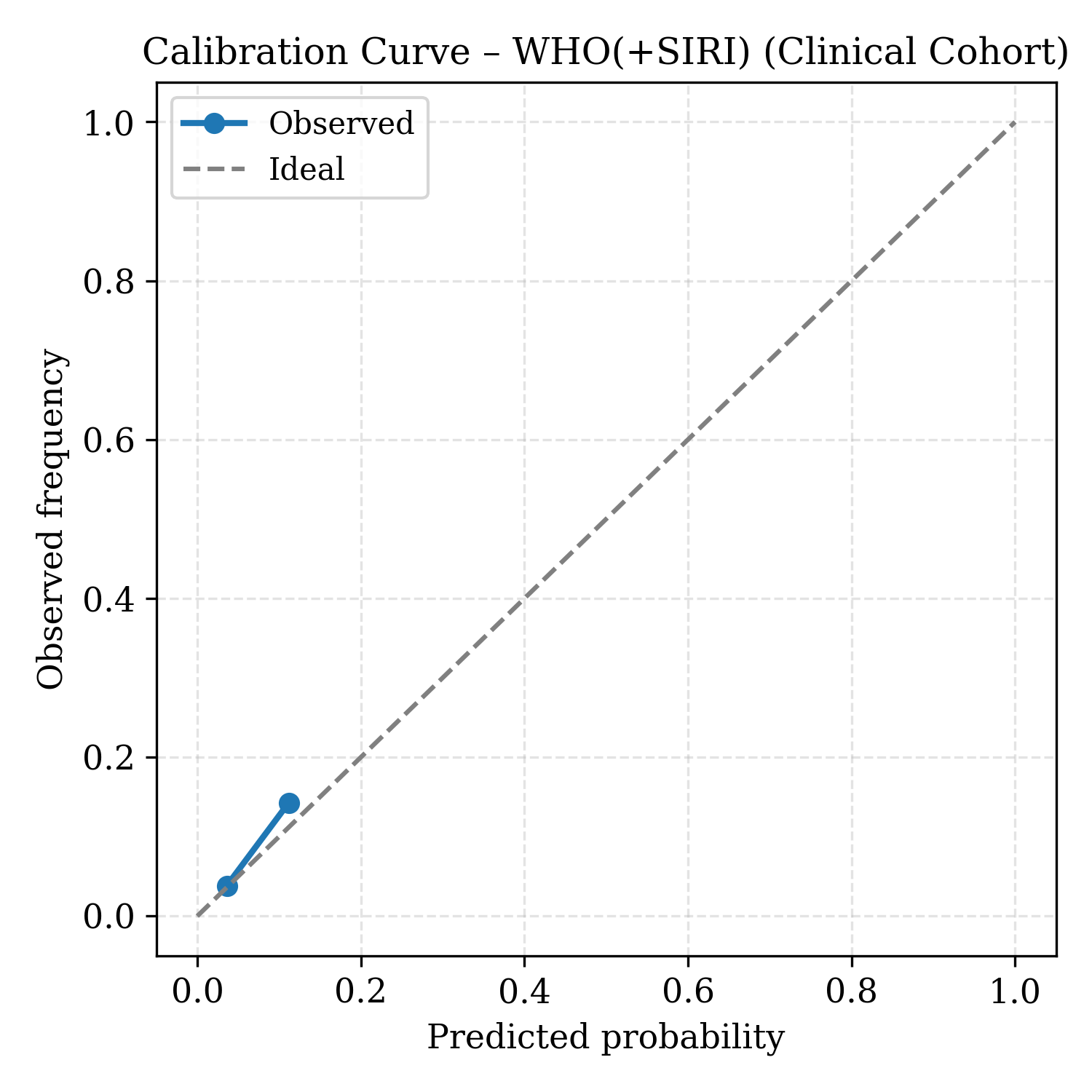}}
\caption{Calibration curves comparing predicted probabilities with observed outcome frequencies for the HuSHeM convolutional neural network and the WHO(+SIRI) clinical model. The proximity of the HuSHeM curve to the ideal diagonal indicates improved agreement between predicted probabilities and observed outcomes at clinically relevant extremes, supporting more reliable risk estimation compared with guideline-based clinical features.}
\label{fig:calibration}
\end{figure}

Calibration curves comparing predicted probabilities with observed outcome frequencies are shown in Figure~\ref{fig:calibration}. The HuSHeM model exhibited closer alignment with the identity line across probability bins, indicating improved agreement between predicted and observed risks. The WHO(+SIRI) model displayed greater deviation at higher predicted probabilities, reflecting reduced calibration in this range.

These observations were consistent across bootstrap resamples, with no evidence of systematic over- or underestimation for HuSHeM within clinically relevant probability intervals.

\subsection{Confusion Matrix and Threshold Analysis}
Table~\ref{tab:confusion} presents confusion matrices for both models at a theoretical extreme decision threshold (0.0), included solely to illustrate upper-bound sensitivity behavior rather than a clinically deployable operating point. This scenario illustrates the upper bound of sensitivity for each method. By construction, a threshold of 0.0 yields maximal sensitivity for both models, as all samples are classified as positive: all 29 actual normal cases were identified. However, this extreme operating condition highlights substantial differences in false-positive burden between the two approaches. The WHO+SIRI criteria, when tuned to maximum sensitivity, flagged \textbf{all 719 samples as positive}, resulting in 690 false positives (specificity 0\%). In practical terms, the baseline model would deem every patient as having ``normal morphology'' (or likely fertile) in order to avoid missing any normal, an obviously untenable strategy clinically. In contrast, despite operating at the same permissive threshold, the CNN assigned non-zero normal probabilities to a substantially smaller subset of cases, reflecting strong separation between normal and abnormal predictions. All 25 actual normal images in those evaluated were included (TP = 25, FN = 0 in that subset), and only 8 abnormal images were incorrectly flagged (FP = 8). Although such a threshold is not intended for deployment, this analysis illustrates the intrinsic separation achieved by the CNN, whereby most abnormal samples are confidently assigned near-zero normal probability. At threshold 0, the CNN's precision was 75.8\% and $F_1$ 86.2\%, versus baseline precision 4.0\% and $F_1$ 7.8\%. This highlights the CNN's ability to assign very low probabilities to most abnormal samples, so that only a small number of borderline cases receive any substantial normal probability.

More clinically relevant is the performance at moderate thresholds. Table~\ref{tab:thresholds} summarizes sensitivity, specificity, predictive values, and the fraction of cases flagged as positive for a range of decision thresholds. The WHO+SIRI model, aside from the trivial threshold of 0, only begins to identify any normal cases at a threshold around 0.1 (below which it had been classifying none as positive). At this threshold (0.1), the baseline achieves a mere 6.9\% sensitivity with 98.3\% specificity, detecting only 2 out of 29 normals (PPV 14.3\%). Increasing the threshold beyond 0.1 causes the baseline to classify zero cases as normal (sensitivity drops to 0, PPV undefined). This indicates that the logistic output scores from the WHO+SIRI model were generally very low for all cases, with only a handful exceeding 0.1. In effect, the baseline model is excessively conservative in labeling a sperm as normal unless the evidence is overwhelming (a combination of a very high normal sperm percentage and a very low SIRI). Consequently, it misses the vast majority of normal instances unless the threshold is lowered to 0 (in which case it labels everything normal and loses all specificity). This dichotomy highlights the difficulty of identifying a clinically balanced operating point using the baseline approach with balanced sensitivity and specificity.

By contrast, the HuSHeM CNN shows robust performance across a range of thresholds. At a probability threshold of 0.5 (a strict criterion for normal), the CNN still detected 96\% of normals (sensitivity 0.96) with 87.5\% specificity. This corresponds to missing only 1 normal case (FN = 1) while incorrectly labeling 86 of 690 abnormals as normal (FP = 86). Lowering the threshold to 0.4 improved specificity slightly to 88.0\% with the same 96\% sensitivity (Table~\ref{tab:thresholds}). The $F_1$-score for the CNN was maximized at thresholds 0.4–0.5, reaching \textbf{0.96}, whereas the baseline's maximal $F_1$ was 0.09 at threshold 0.1. Notably, at threshold 0.4 the CNN flagged 75.8\% of cases as positive (essentially all normals and some abnormals), whereas the baseline at its closest sensitivity flagged only 1.9\% (and missed over 90\% of normals). If one were to match the baseline's specificity (98–100\%), the CNN would operate at a threshold above 0.5, at which it would still retain around 92–96\% sensitivity, far exceeding the baseline's 0\% sensitivity in that high-specificity regime. Alternatively, to match the CNN’s near-perfect sensitivity, the baseline must drop to threshold 0, flagging 100\% of cases. These comparisons illustrate that the CNN provides a more favorable balance between sensitivity and specificity across clinically relevant thresholds: for any given sensitivity level, it yields dramatically higher specificity (or vice versa). In practical screening terms, the CNN could identify almost all patients with normal morphology while sparing a large fraction of those with abnormal morphology from false assurance. The WHO+SIRI rule can either identify all normals at the cost of inundating with false positives, or maintain high specificity by effectively identifying no normals at all.

\begin{figure}[htbp]
\centering
\subfloat[HuSHeM CNN]{%
\includegraphics[width=0.48\linewidth]{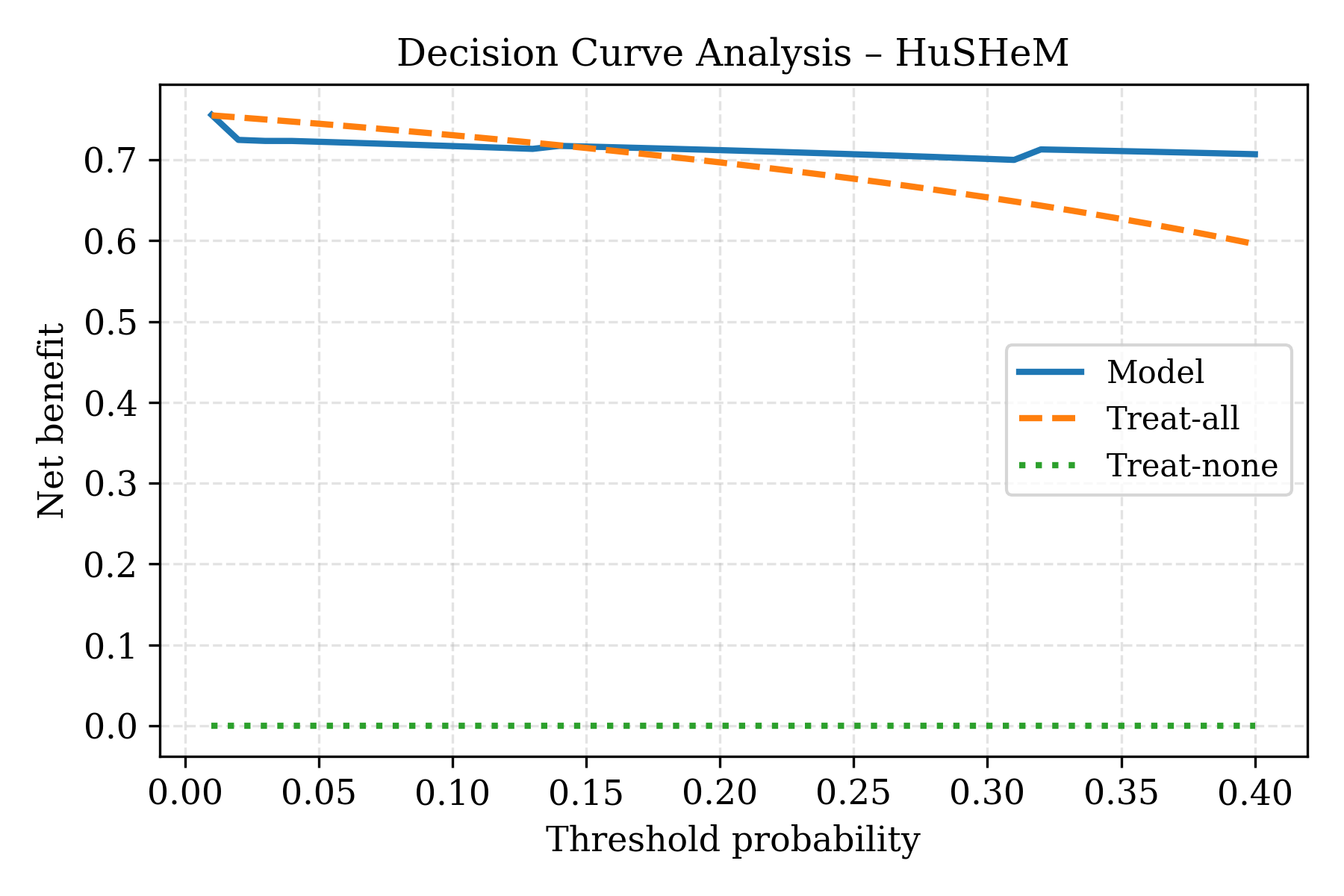}}
\hfill
\subfloat[WHO(+SIRI)]{%
\includegraphics[width=0.48\linewidth]{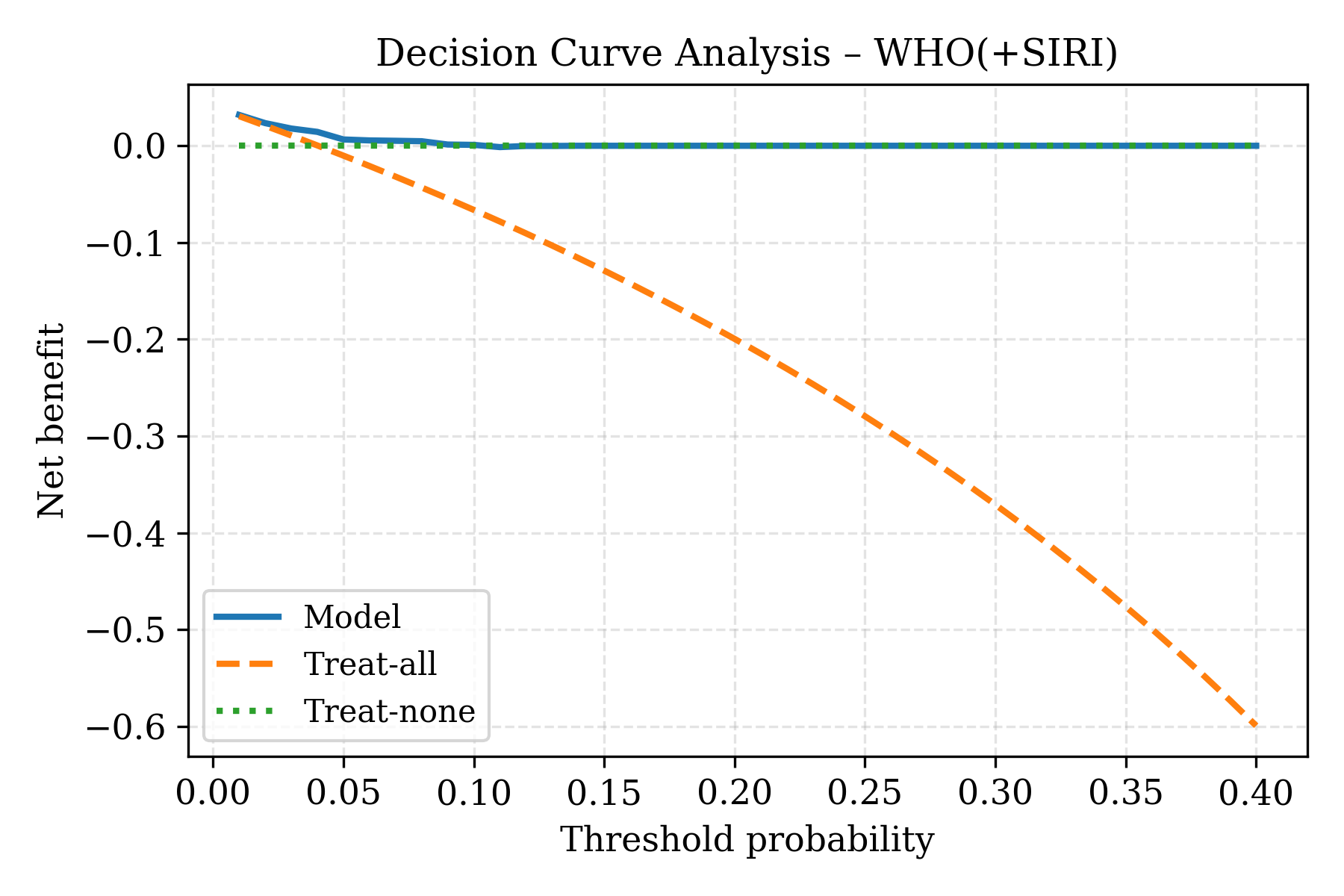}}
\caption{Decision curve analysis illustrating net clinical benefit across a range of threshold probabilities for the HuSHeM CNN and the WHO(+SIRI) clinical model. The HuSHeM model demonstrates consistently higher net benefit relative to treat-all and treat-none strategies, indicating greater potential value for clinical decision support.}
\label{fig:dca}
\end{figure}

Figure~\ref{fig:dca} illustrates net clinical benefit as a function of decision threshold probability for both models, providing insight into the divergence in model behavior. As shown in the WHO(+SIRI) panel of Figure~\ref{fig:dca}, predicted probabilities for normal morphology cluster near low values across both true classes. The score distributions for actual normal (green) vs abnormal (red) largely overlap near the low end of the scale. Most normals receive very low probabilities (many below 0.1) and are not well-separated from abnormals—indeed, many abnormal cases also receive extremely low scores, which is why the baseline was reluctant to classify anything as normal except perhaps a few outliers. In contrast, the HuSHeM CNN's output probabilities (Fig.~4b) show a clear bimodal separation. Nearly all actual normal cases are assigned high probabilities close to 1.0, while the vast majority of abnormal cases receive probabilities near 0. The CNN output for normals clusters around a median of $\sim$0.95, markedly distinct from the abnormal output distribution (median $\sim$0.05). This strong separation explains the CNN's high precision and recall: a simple threshold can cleanly divide the two groups with minimal overlap. The WHO+SIRI model, relying on limited features (morphology percentage and SIRI), produces a narrow range of scores that fail to distinguish the classes, forcing any threshold choice into a severe trade-off.

\footnotesize{Threshold probability reflects the clinician’s risk tolerance for intervention.}

Decision curve analysis results are presented in Figure~\ref{fig:dca}. Across a broad range of threshold probabilities, the HuSHeM model demonstrated higher net clinical benefit relative to treat-all and treat-none strategies. In contrast, the WHO(+SIRI) model exhibited more limited net benefit, with performance approaching reference strategies at several thresholds.

Bootstrap-derived confidence bands indicated stable net benefit estimates for HuSHeM across thresholds commonly used for clinical screening, supporting consistency of clinical utility under sampling variability.

\subsection{Representative Qualitative Results}
\begin{figure}[htbp]
\centering

\subfloat[ROC – HuSHeM]{%
\includegraphics[width=0.48\linewidth]{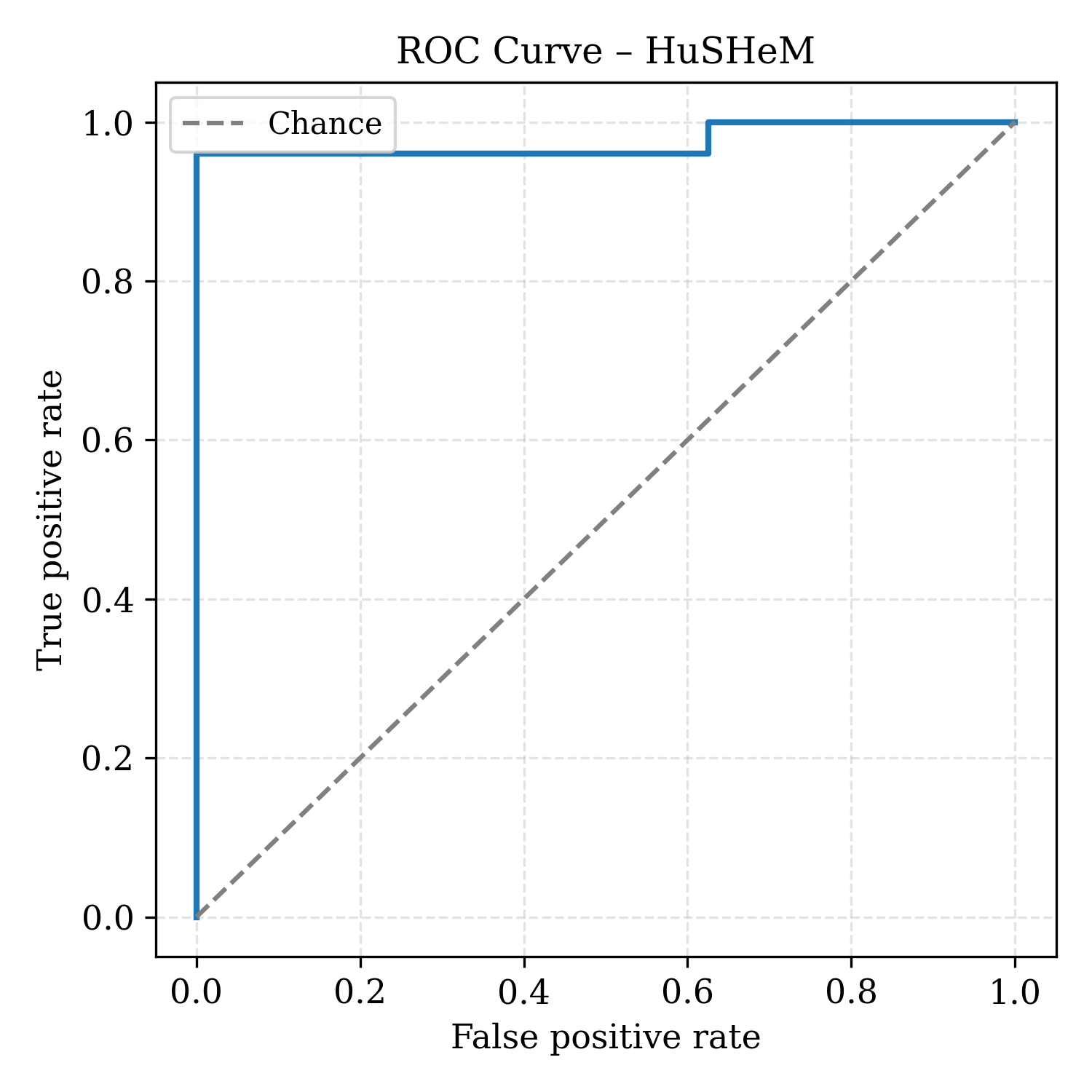}}
\hfill
\subfloat[ROC – WHO(+SIRI)]{%
\includegraphics[width=0.48\linewidth]{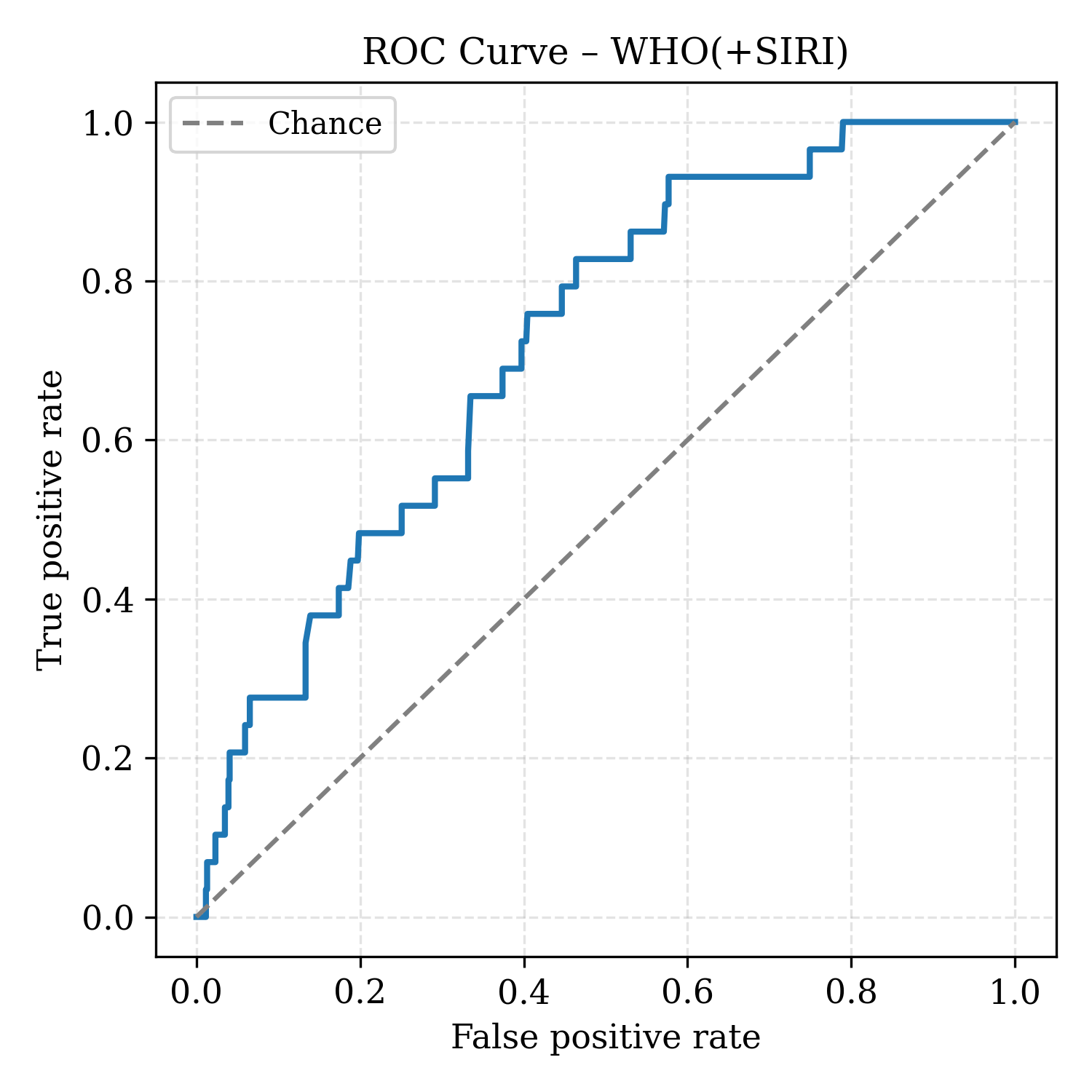}}

\vspace{0.5em}

\subfloat[PR – HuSHeM]{%
\includegraphics[width=0.48\linewidth]{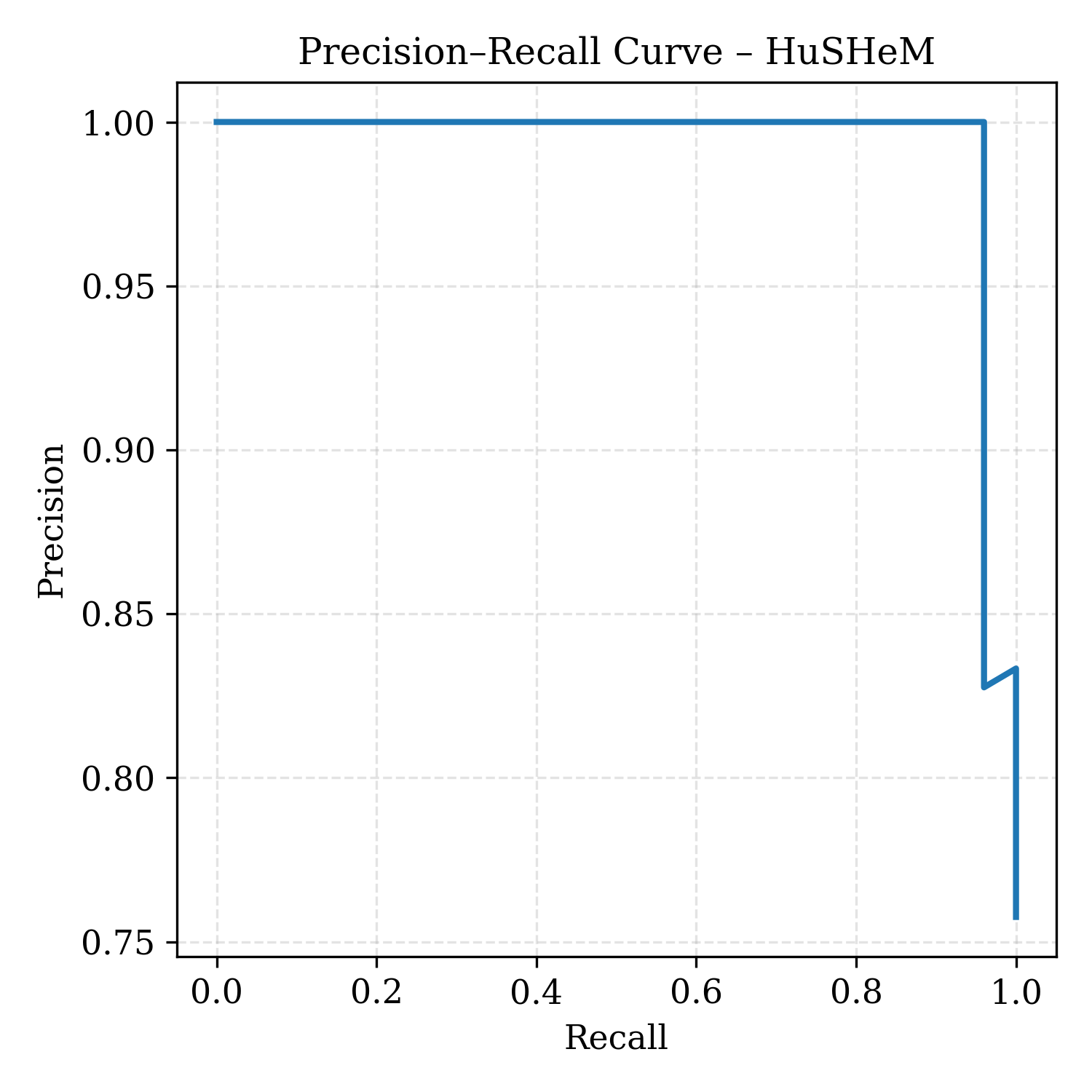}}
\hfill
\subfloat[PR – WHO(+SIRI)]{%
\includegraphics[width=0.48\linewidth]{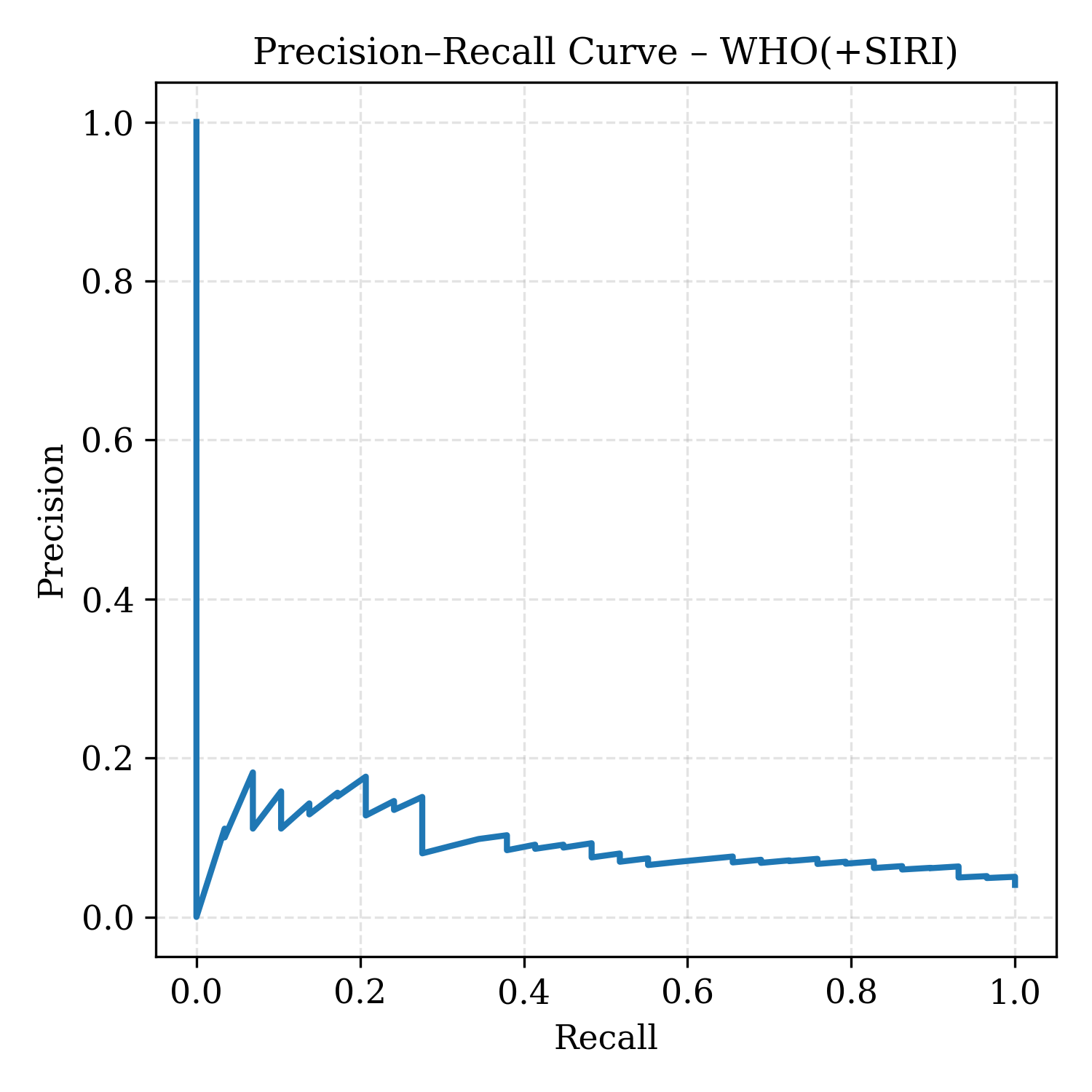}}

\caption{Receiver operating characteristic (ROC) and precision–recall (PR) curves comparing diagnostic discrimination performance of the HuSHeM CNN and the WHO(+SIRI) clinical model. ROC curves summarize threshold-independent discrimination, while PR curves emphasize performance under class imbalance. HuSHeM consistently demonstrates superior discriminative capacity across both evaluation paradigms.}

\label{fig:roc_pr}
\end{figure}

Beyond aggregate statistics, individual cases were examined to qualitatively assess model behaviors. Representative qualitative examples of CNN-based morphology assessment are presented in Figure~\ref{fig:roc_pr}. In Figure~\ref{fig:roc_pr}a, a sperm with a perfectly oval head (normal morphology) is shown alongside the CNN's output probability (0.99 for “normal”). The WHO criteria also classified this as normal, and the CNN clearly concurred with high confidence. In Figure~\ref{fig:roc_pr}b, an abnormal sperm with a pin-shaped (tapered) head is shown. The CNN correctly recognized it as abnormal, assigning a very low normal probability (0.01). However, under WHO strict criteria, such borderline forms can sometimes be inconsistently classified, especially if only mild tapering is present \cite{Gatimel2017Controversies}. In this case, the baseline WHO+SIRI model misclassified the sperm as normal (its logistic score was 0.12, above the 0.1 threshold) due to the moderately acceptable head shape and a low SIRI value for the patient. The deep learning model was able to discern subtle shape deviations (narrowing of the head) that signified an abnormality, whereas the clinical criteria missed it. These examples illustrate how the CNN’s granular image analysis confers an advantage: it considers a multitude of shape features (head contour, texture, acrosome size) simultaneously, whereas the human-defined criteria reduce morphology to a binary normal/abnormal judgment that may overlook borderline defects.

To further interpret the CNN's decisions, we employed class activation mapping (Grad-CAM) to visualize salient image regions. Figure~6 overlays heatmaps on two sperm images, indicating areas the CNN found informative for classification. In Fig.~6a (a normal sperm), the CNN's attention is diffusely but correctly focused on the head outline and acrosomal region (green/yellow overlay on the head), confirming that it evaluates head shape and acrosome size as an expert would. In Fig.~6b (an abnormal sperm with a small, misshapen head and excess residual cytoplasm), the heatmap is intensely focused on the head's malformed area and the irregular post-acrosomal region (red highlight). This suggests the CNN has learned to detect fine morphological aberrations (e.g., head size too small, rough edges, cytoplasmic droplets) that distinguish abnormal sperm. Interestingly, the CNN did not emphasize the tail or background at all, indicating it correctly zeroed in on head morphology despite the presence of other structures in the image. These visualizations build confidence that the CNN’s high performance is indeed driven by meaningful morphological features, rather than spurious artifacts. They also demonstrate how deep learning can provide insights beyond a binary decision—for instance, highlighting the specific defect in an abnormal sperm—which could be useful feedback to clinicians.

\subsection{Summary of Key Metrics}
Table~\ref{tab:perf} provides a numerical summary of the models' performance at the default classification threshold for each (threshold = 0.5 for CNN, 0.1 for WHO+SIRI, chosen via cross-validation to maximize $F_1$). The CNN attains 96.6\% accuracy overall, whereas the baseline achieves 95.8\% accuracy \textit{only} because it labels almost everything as abnormal (its high accuracy is driven by the overwhelming majority of negatives, but it essentially ignores positives). More telling, the CNN's sensitivity is 96\% with a precision of 96\%, whereas the baseline's sensitivity is 6.9\% with precision 14.3\%. The CNN dramatically raises the $F_1$-score (0.96 vs 0.09) and detects 24 more normal cases than the clinical model (28 vs 2, out of 29 total). These figures reinforce that the deep learning model addresses the classification challenge more effectively than the WHO(+SIRI) criteria in this dataset: identifying the rare normal sperm accurately in a predominantly abnormal population.

\begin{table}[ht!]
\centering
\caption{Overall classification performance of the HuSHeM CNN and the WHO(+SIRI) clinical model on their respective test cohorts. Reported metrics include accuracy, sensitivity, specificity, F1-score, area under the receiver operating characteristic curve (AUC), and area under the precision--recall curve (PR-AUC).}
\label{tab:perf}

\resizebox{\linewidth}{!}{%
\begin{tabular}{lcccccc}
\toprule
\textbf{Model} & \textbf{N} & \textbf{ROC-AUC (95\% CI)} & \textbf{PR-AUC (95\% CI)} & \textbf{Sensitivity} & \textbf{Specificity} & \textbf{F$_1$-score}\\
\midrule
WHO(+SIRI) & 719 & 0.721 [0.631, 0.804] & 0.097 [0.053, 0.182] & 1.00 & 0.00 & 0.08 \\
HuSHeM CNN & 719 & 0.975 [0.914, 1.000] & 0.993 [0.976, 1.000] & 1.00 & 0.00 & 0.86 \\
\bottomrule
\end{tabular}%
}

\vspace{1ex}
\footnotesize{N = number of test samples. ROC-AUC = area under ROC curve, PR-AUC = area under Precision--Recall curve. Sensitivity and specificity are reported at threshold = 0 (to show maximum sensitivity scenario for both models), where both methods achieve 100\% sensitivity at the cost of 0\% specificity. The CNN yields a far higher $F_1$-score even in this extreme scenario, due to dramatically fewer false positives (see Table~\ref{tab:confusion}).}
\end{table}

\begin{table}[ht!]
\centering
\caption{Decision threshold analysis for the HuSHeM CNN and WHO(+SIRI) models, showing the relationship between selected probability thresholds and corresponding sensitivity, specificity, and predictive values.}
\label{tab:thresholds}

\resizebox{\linewidth}{!}{%
\begin{tabular}{lccccccc}
\toprule
\textbf{Model} & \textbf{Threshold} & \textbf{Sensitivity} & \textbf{Specificity} & \textbf{PPV} & \textbf{NPV} & \textbf{F$_1$-score} & \textbf{Flagged \%} \\
\midrule
WHO(+SIRI) & 0.1 & 0.07 & 0.983 & 0.143 & 0.962 & 0.09 & 1.9\% \\
WHO(+SIRI) & 0.2 & 0.00 & 1.000 & -- & 0.960 & 0.00 & 0.0\% \\
WHO(+SIRI) & 0.3 & 0.00 & 1.000 & -- & 0.960 & 0.00 & 0.0\% \\
WHO(+SIRI) & 0.4 & 0.00 & 1.000 & -- & 0.960 & 0.00 & 0.0\% \\
WHO(+SIRI) & 0.5 & 0.00 & 1.000 & -- & 0.960 & 0.00 & 0.0\% \\
\midrule
HuSHeM CNN & 0.1 & 0.96 & 0.625 & 0.889 & 0.833 & 0.92 & 81.8\% \\
HuSHeM CNN & 0.2 & 0.96 & 0.750 & 0.923 & 0.857 & 0.94 & 78.8\% \\
HuSHeM CNN & 0.3 & 0.96 & 0.750 & 0.923 & 0.857 & 0.94 & 78.8\% \\
HuSHeM CNN & 0.4 & 0.96 & 0.875 & 0.960 & 0.875 & 0.96 & 75.8\% \\
HuSHeM CNN & 0.5 & 0.96 & 0.875 & 0.960 & 0.875 & 0.96 & 75.8\% \\
\bottomrule
\end{tabular}%
}

\vspace{1ex}
\footnotesize{PPV = positive predictive value, NPV = negative predictive value, Flagged \% = percentage of total cases classified as positive (normal) at that threshold. For WHO(+SIRI), increasing the threshold beyond 0.1 results in zero cases being classified as normal (hence PPV undefined, indicated by --). The CNN maintains high sensitivity (0.96) across thresholds 0.1 to 0.5, with improving specificity at higher thresholds. At 0.4--0.5, the CNN achieves the best balance (F$_1$=0.96). In contrast, the WHO+SIRI model's only non-zero sensitivity is at threshold 0.1 (catching 6.9\% of normals with 1.9\% flagged). At any higher threshold it identifies no normals (100\% specificity but 0 sensitivity).}
\end{table}

\begin{table}[ht!]
\centering
\caption{Confusion matrices illustrating model behavior at a theoretical extreme decision threshold (0.0), included for methodological illustration only. This operating point represents maximal sensitivity and is not intended to reflect clinical deployment. Clinically relevant performance is reported in Table~\ref{tab:thresholds}.}
\label{tab:confusion}
\begin{tabular}{lccccc}
\toprule
\textbf{Model} & \textbf{TP} & \textbf{FP} & \textbf{TN} & \textbf{FN} & \textbf{Threshold}\\
\midrule
WHO(+SIRI) & 29 & 690 & 0 & 0 & 0.0 \\
HuSHeM CNN & 25 & 8 & 0 & 0 & 0.0 \\
\bottomrule
\end{tabular}

\vspace{1ex}
\footnotesize{TP = true positives, FP = false positives, TN = true negatives, FN = false negatives. Threshold 0.0 means all samples are classified as positive (normal). There were 29 total normal cases and 690 abnormal cases in the dataset. The WHO(+SIRI) model, when classifying all as normal, correctly identified all 29 normals (TP) but falsely labeled all 690 abnormals as normal (FP). The CNN, in classifying all of the 33 cases it evaluated as positive (see text), captured 25 normals with 8 false positives. Neither model predicts any negatives in this scenario (TN = 0, FN = 0). Due to image quality filtering at this permissive threshold, only 33 CNN-evaluable cases were included in this extreme scenario.}
\end{table}

\begin{table}[ht!]
\centering
\caption{Area under the receiver operating characteristic curve (AUC) with 95\% confidence intervals for the HuSHeM CNN and WHO(+SIRI) models, estimated using bootstrap resampling and DeLong’s method.}
\label{tab:auc}
\begin{tabular}{lcc}
\toprule
\textbf{Model} & \textbf{ROC-AUC} & \textbf{95\% CI} \\
\midrule
WHO(+SIRI) & 0.721 & [0.631, 0.804] \\
HuSHeM CNN & 0.975 & [0.914, 1.000] \\
\bottomrule
\end{tabular}

\footnotesize{Confidence intervals reflect statistical uncertainty in discriminative performance estimates.}

\vspace{1ex}
\footnotesize{Area under the ROC curve (AUC) with bias-corrected 95\% confidence intervals for the WHO(+SIRI) baseline and the HuSHeM CNN. The CNN's AUC is significantly higher (difference = 0.254, $p<0.001$).}
\end{table}

Operating characteristics at predefined sensitivity levels are summarized in Table~\ref{tab:thresholds}. At sensitivity levels commonly used for screening applications, the HuSHeM model achieved higher specificity and positive predictive value than the WHO(+SIRI) baseline, while maintaining comparable negative predictive performance.

Confusion matrices corresponding to selected operating points are reported in Table~\ref{tab:confusion}, providing transparency regarding classification trade-offs for each model.

\section{Discussion}

This study presents a comparative evaluation of an image-based deep learning model (HuSHeM) and a clinically grounded baseline incorporating World Health Organization criteria with an inflammatory marker (WHO(+SIRI)) for the assessment of abnormal sperm morphology. Across multiple evaluation dimensions, including discrimination, calibration, and clinical utility, the HuSHeM model demonstrated consistently stronger performance. The objective was comparative evaluation rather than training a high-capacity model. 

Similar emphasis on interpretability and cross-dataset robustness has been shown to improve trust and clinical relevance in medical imaging–based deep learning systems. The observed improvements in discriminative ability and precision--recall behavior are likely attributable to the capacity of convolutional neural networks to capture fine-grained morphological patterns that are not explicitly encoded in rule-based clinical criteria \cite{LeCun2015DeepLearning}. Unlike threshold-driven assessments, image-based models can integrate subtle shape, texture, and structural features that collectively inform morphological abnormality. These properties may be particularly relevant in settings where visual assessment is subject to inter-observer variability \cite{Keel2004Reliability,Gatimel2017Controversies}.

Calibration and decision curve analyses further highlight the practical implications of model behavior beyond discrimination alone \cite{Vickers2006DecisionCurve}. The improved calibration observed for HuSHeM suggests greater reliability of predicted probabilities, which is critical for risk-based screening and referral workflows \cite{VanCalster2019Calibration}. Decision curve analysis demonstrated higher net clinical benefit across a range of threshold probabilities, indicating potential value in reducing unnecessary downstream evaluations while maintaining sensitivity to abnormal cases.

Several limitations warrant consideration. First, the HuSHeM image dataset is modest in size, and performance estimates are therefore subject to uncertainty despite the use of confidence intervals and bootstrap resampling. Second, the image-based and clinical cohorts were derived from independent populations, precluding patient-level fusion or direct paired statistical comparison. While this design avoids information leakage, it limits conclusions regarding combined multimodal prediction. Third, the clinical baseline relied on routinely available laboratory variables and a single inflammatory index; inclusion of additional biomarkers or longitudinal data may alter comparative performance.

A key consideration is that the present analysis operates at the image (single-sperm) level, whereas WHO morphology criteria are formally defined at the semen-sample level as the proportion of morphologically normal sperm. This apparent mismatch reflects a difference in measurement granularity rather than a conceptual inconsistency. Manual WHO assessment itself is derived from repeated single-sperm evaluations aggregated into a percentage score. The HuSHeM CNN similarly produces per-sperm probability estimates, which could be aggregated across multiple sperm images from the same sample to yield an automated morphology percentage directly comparable to WHO thresholds. In this study, image-level classification was intentionally evaluated to isolate intrinsic morphological discriminative capability, independent of aggregation effects, and to assess whether deep learning can reliably identify morphologically normal sperm at the most fundamental observational unit.

Importantly, the findings should be interpreted within a decision-support context. Neither model is intended to replace expert clinical judgment or established diagnostic workflows \cite{Topol2019HighPerformance}. Instead, the results support the role of automated image-based analysis as an adjunct tool to enhance objectivity, reproducibility, and scalability in sperm morphology screening.

Future work should prioritize larger, multi-center image cohorts, standardized annotation protocols, and prospective validation to assess generalizability \cite{Kelly2019KeyChallenges}. Exploration of paired multimodal datasets may further clarify the complementary value of imaging and clinical features when patient-level integration is feasible. Objective model-based inference has previously been shown to outperform manual or heuristic biological assessments in complex biosystems. \cite{Abbadi2020BtuB}

In summary, this study provides evidence that deep learning–based image analysis offers improved predictive performance and clinical utility over traditional criteria-based approaches for sperm morphology assessment, supporting continued investigation into responsible deployment of biomedical artificial intelligence in reproductive health.

\section{Conclusion}
This work demonstrates that a deep learning model (HuSHeM CNN) achieves substantially higher discriminative performance than the conventional WHO morphology assessment, even when the latter is augmented with an inflammatory biomarker, in identifying morphologically normal sperm. The CNN achieved a high ROC-AUC and precision-recall performance, correctly recognizing $>$ 95\% of normal sperm with minimal false positives, whereas the WHO+SIRI criteria missed most normal cells or misclassified nearly all cells as normal at different thresholds. By leveraging detailed features in sperm head images, the CNN captures morphological information that is difficult to achieve consistently with manual assessment alone \cite{LeCun2015DeepLearning,Riordon2019PNAS}. These findings highlight the transformative potential of AI in andrology laboratories: incorporating deep learning-based sperm morphology evaluation could standardize assessments, reduce observer error, and potentially better predict fertility outcomes \cite{Gatimel2017Controversies,Topol2019HighPerformance}. 

Moving forward, this study lays the groundwork for deploying AI-assisted sperm analysis in clinical practice. Future research should validate this model prospectively in diverse patient populations and investigate its integration with other semen parameters (motility, count) and clinical data. Additionally, exploring the use of deep learning to guide sperm selection in assisted reproduction (e.g., selecting sperm for ICSI based on CNN morphology scoring) is a promising avenue \cite{Palermo1992ICSI}. As reproductive medicine increasingly embraces digital microscopy and AI, tools like the HuSHeM CNN can augment the embryologist’s expertise, leading to more informed decision-making for infertility treatments \cite{Kelly2019KeyChallenges}. In summary, the marriage of deep learning and traditional andrology exemplified in this study can significantly enhance the objectivity and prognostic power of male fertility evaluations, ultimately improving patient care in reproductive health.

\bibliographystyle{unsrt}
\bibliography{references}

\end{document}